\documentclass[accepted,specialissue]{melba}

\usepackage{mwe} 

\usepackage{amsmath,amsfonts}

\usepackage{lipsum}
\usepackage{subcaption}
\usepackage{gensymb}
\usepackage{tikz}
\usepackage[most]{tcolorbox}
\newtcolorbox{highlightbox}{colback=yellow!25, colframe=yellow!25, boxrule=0pt, arc=2pt}
\usepackage{xcolor}
\usepackage{soul}
\sethlcolor{yellow} 

\melbaid{2026:016}  
\doi{10.59275/j.melba.2026-522e}
\melbaauthors{Kubík, Guibault, Španěl and Lombaert}  
\email{ikubik@fit.vut.cz}
\volume{2026}
\firstpageno{313}  
\melbayear{2026}  
\datesubmitted{2025-10-31}  
\datepublished{2026-06-08}  

\melbaspecialissue{Information Processing in Medical Imaging (IPMI) 2025}
\melbaspecialissueeditors{Ipek Oguz, Shireen Elhabian, Ismail Ben Ayed}

\ShortHeadings{Deep Spectral Models for Robust Dental Shape Generation}{Kubík \textit{et al.}}

\title{Deep Spectral Models for Robust Dental Shape Generation}

\author{
	\firstname Tibor \surname Kubík\aff{1,2},
	\name François Guibault\aff{1},
    \name Michal Španěl\aff{2},
    \name Hervé Lombaert\aff{1},
}
\affiliations{
	\num 1 \addr Polytechnique Montréal, Montréal, Canada \\
	\num 2 \addr Brno University of Technology, Brno, Czech Republic
}

\abstract{
	Accurate modeling of dental crown morphology is fundamental for diagnosis, orthodontic planning, and computer-aided restoration design.
    However, datasets suitable for training such models are typically limited in size.
    We present ToothForge, a deep spectral generative framework that models dental crown geometries from compact, intrinsic representations.
    By operating in the spectral domain, ToothForge learns a latent manifold of 3D tooth shapes through synchronized spectral embeddings, ensuring consistent modeling across samples with varying connectivity.
    Spectral synchronization mitigates the instability of Laplace-Beltrami eigenbases and enables efficient learning in a low-dimensional space.
    The framework is thoroughly evaluated through robustness analysis, ablation studies, and benchmarking against PCA-based statistical shape models and point-based generative frameworks.
    Results show that synchronized spectral modeling achieves reconstruction and generative performance comparable to or exceeding spatial approaches, while maintaining compactness and geometric interpretability.
    Together, the compact synchronized coefficients and low-dimensional learning space make the framework particularly suitable for limited datasets, as often encountered in dental and medical domains, and applicable in real-world scenarios where guaranteeing consistent connectivity across shapes from various clinics is unrealistic.
}

\keywords{3D Tooth Shape Generation, Digital Dentistry, Spectral Shape Learning, Geometric Deep Learning}

\begin{document}

\twocolumn[\maketitle]


\section{Introduction}
Digital dentistry relies heavily on 3D shape analysis for tasks such as morphology assessment and computer-aided dental restorations.
Among morphology assessment applications, accurate modeling of tooth and root morphology plays a critical role in several clinical domains.
In dentofacial orthopaedics, evaluating palatal shape morphology has the potential to aid in the evaluation and outcome prediction of maxillary expansion~(\citeauthor{matsuyama:palatal-clinical-ii,nauwelaers:palatal-variation-dl,primovzivc:palatal-clinical}). 
In endodontics, dental pulp treatment, knowledge of root canal anatomy is crucial for guiding canal detection and preventing procedural complications such as perforation~(\citeauthor{cleghorn:endo-clinical-i,peiris:endo-clinical-ii}).
Similarly, in implantology, knowledge of crown and root morphology determines implant trajectory, proximity to the sinus floor, and the feasibility of immediate implant placement after extraction~(\citeauthor{bhola:implants-clinical-i,testori:implants-clinical-ii}).

Beyond diagnostic and surgical applications, automated generation of dental restorations represents key domain where accurate 3D modeling of tooth morphology is critical~(\citeauthor{kong:ai-prosthesis}).
In restorative dentistry, digital workflows increasingly rely on data-driven crown design, where prosthetic shapes must adapt to the patient’s existing anatomy and occlusal context. 
The ability to synthesize realistic and anatomically feasible crown restorations can substantially reduce manual design time and thus improve clinical and laboratory practice~(\citeauthor{golriz:crown-generation,tian:crown-generation,yang:crown-generation}).
However, these frameworks require large, diverse, and well-annotated datasets of dental shapes to achieve generalization across patients and tooth classes. 
Such requirement is rarely met in practise due to privacy constraints and costly annotations.

Collectively, these applications highlight two complementary needs in digital dentistry:
(i) shape models that capture population-level variability of dental morphology for analysis and diagnosis, and
(ii) generative frameworks capable of synthesizing realistic shapes for restorative design and data augmentation.

As a traditional baseline, statistical shape models (SSMs) model shape variation by principal component analysis (PCA) of training data shapes~(\citeauthor{wold:pca}). 
By representing a population of shapes through low-dimensional linear modes of variation, PCA-based SSMs provide interpretable and efficient frameworks for morphological analysis and reconstruction.
They also possess generative power by modifying the variables within given shape population.
However, due to their inherently linear formulation, these models struggle to capture complex anatomical geometries. 
Examples include the non-linear evolution of palatal dimensions during dentition development and the non-convex relationships among occlusal surfaces and cusp morphologies.

The recent emergence of 3D deep learning has extended the paradigm of SSMs toward non-linear representations, usually operating over spatial coordinates (\citeauthor{adams:pcssms,pan:variational-point-ae,yang:pointflow,yang:foldingnet, becker:point-deep-models,luo:point-diffusion}).
Autoencoder (AE) networks can be described as a non-linear generalization of PCA. They are capable of capturing population-level variability while serving as a generative model for synthetizing novel shapes, both in a non-linear setting.
Moreover, AEs are easily expanded to provide unified modeling across multiple classes, integrate contextual or multimodal inputs, or perform additional downstream analysis like classification, regression or pathology detection.
Despite their flexibility, autoencoders also introduce several practical limitations when applied to anatomical shape modeling.
First, they scale poorly with resolution. 
Training directly on high-dimensional spatial coordinates or dense meshes requires large datasets and considerable computational resources.
This is particularly challenging in medical domains such as digital dentistry, where datasets are small but each sample contains tens or hundreds of thousands of vertices~(\citeauthor{golriz:dental-dataset-4,kubik:dental-dataset-1,tan:dental-dataset-3,wang:dental-dataset-2}).
This dimensionality leads to a severe imbalance between data size and representational dimensionality.
In such settings, overfitting becomes a critical risk, and training stability often depends on extensive downsampling, which may compromise geometric fidelity.
Second, unlike PCA-based SSMs, the latent representations learned by AEs are not inherently orthogonal or ordered by explained variance.
As a result, latent dimensions can be entangled, making it difficult to attribute specific anatomical meaning to individual dimensions without additional constraints~(\citeauthor{chen:vae-disentanglement}).

To address these challenges, recent research~(\citeauthor{reuter:shapeDNA,reuter:discreteLBO,lemeunier:specae,lemeunier:spectrans,biffi:explainable-ae}) has explored spectral representations of 3D surfaces as a compact lternative to spatial coordinates.
By decomposing a surface into frequency coefficients of the Laplace-Beltrami operator, shapes can be described through intrinsic, ordered features, where low frequencies capture global form and higher modes encode fine details.
Such a decomposition facilitates more efficient learning while maintaining geometric interpretability.
The applicability of spectral decomposition to deep learning was demonstrated by \cite{lemeunier:specae}, who trained an autoencoder on truncated Laplacian coefficients and showed that networks can learn geometry in the spectral domain.
However, their approach requires all training meshes to share identical connectivity, making it unsuitable for clinical datasets with heterogeneous mesh connectivity and varying vertex counts. A similar requirement for known point-wise correspondences is present in recent point-based diffusion models, such as that of ~\cite{zhu:ptdiff-correspondences}.
In earlier preliminary work~(\citeauthor{kubik:toothforge}), we explored this limitation by introducing spectral synchronization, which aligns the spectral embeddings of all shapes to a common reference basis.
This alignment removes the bias introduced by eigenbasis instability and enables modeling of a shared spectral manifold across arbitrary connectivities.

\subsection{Contributions}
The present work develops a complete and rigorously evaluated framework for dental shape generation based on synchronized spectral embeddings.

In summary, the major contributions of our work are as follows:
\begin{itemize}
    \item A deep generative framework for dental shape analysis operating on compact, synchronized spectral coefficients trainable on datasets with variable triangulations.
    \item Evidence that spectral synchronization is essential for achieving high-fidelity reconstructions and that latent regularization performed entirely in the spectral domain is sufficient for effective learning. 
    Optimization in spatial coordinates provides no significant benefit, reinforcing the hypothesis that intrinsic spectral representations are sufficient to model anatomical variability.
    \item A comprehensive experimental evaluation demonstrating the robustness of spectral synchronization with respect to the reference shape, number of modes, and tooth class. 
    The results confirm that the proposed framework generalizes across different tooth families and remains effective under varying spectral truncations, enabling modeling at multiple levels of geometric detail.
    \item  Direct benchmarks against PCA-based statistical shape models and point-based VAEs and diffusion frameworks, supported by a thorough discussion of the observed results and implications for future extensions.
\end{itemize}

This study builds upon and significantly extends the preliminary version presented at the 29th International Conference on Information Processing in Medical Imaging, IPMI 2025~(\citeauthor{kubik:toothforge}).
Beyond expanding the experimental design across multiple tooth classes and truncation levels, the present work adds ablation analyses isolating the roles of synchronization and latent regularization, and introduces comprehensive comparative evaluations against established baselines.
Together, these contributions advance well beyond the preliminary study, establishing a rigorous understanding of the behavior, clinical motivation, and more details of the evaluated methods.
Access the codebase here: \href{https://github.com/tiborkubik/toothForge}{https://github.com/tiborkubik/toothForge}.

\section{Materials and Methods}
This study benchmarks four approaches to generative modeling of dental shapes, concentrating on individual crown geometries without incorporating root structures or anatomical context.
A classical PCA-based statistical shape model provides a linear baseline that is compact and interpretable but limited in expressiveness. 
Second, we consider spatial deep generative models operating directly on point clouds, which overcome linearity by exploiting nonlinear neural mappings but remain constrained by the unordered and extrinsic nature of point sets. 
Finally, we describe \emph{ToothForge}, a spectral generative framework that leverages intrinsic geometry to achieve compact, stable, and connectivity-independent modeling. 
The following subsections introduce each method in detail, followed by the dataset and evaluation metrics description.

\subsection{PCA: Statistical Shape Model}
Principal component analysis (PCA) provides the foundation of classical statistical shape modeling (SSM). 
The main idea is to capture population variability by projecting shapes onto a low-dimensional linear subspace spanned by orthogonal modes of variation.   
SSM is constructed by first bringing all meshes of a given class into
dense correspondence, ensuring that each vertex across the population encodes the same anatomical landmark. 
Each shape with $n$ vertices is then represented as a column vector $x \in \mathbb{R}^{3n}$ obtained by concatenating its $x,y,z$ coordinates. 
Collecting $m$ such shapes yields a data matrix $X = [x_1, \ldots, x_m] \in \mathbb{R}^{3n \times m}$, which is centered by subtracting the mean shape
\[
\bar{x} = \frac{1}{m} \sum_{j=1}^m x_j.
\]
PCA is performed on the covariance matrix of $X$, yielding eigenvectors $U = [u_1, \ldots, u_r]$ and eigenvalues $\lambda_1 \geq \ldots \geq \lambda_r$, where $u_i$ defines the $i$-th principal mode of deformation.  
Any shape $x$ can be approximated in this basis as
\[
x \approx \bar{x} + U_r b,
\]
where $U_r \in \mathbb{R}^{3n \times r}$ contains the leading $r$ eigenvectors and $b \in \mathbb{R}^r$ are the shape parameters. 
The variance explained by each mode is determined by $\lambda_i$, and the coefficients follow a Gaussian distribution $b_i \sim \mathcal{N}(0, \lambda_i)$.  

New shapes can be synthesized by sampling coefficients from the Gaussian prior and reconstructing
\[
x_{\text{new}} = \bar{x} + U_r b.
\]
This linear model is compact, interpretable, and data-efficient. Each eigenvector corresponds to a distinct mode of anatomical variability, and varying $b_i$ allows controlled exploration of the associated deformation.  

However, this linear modeling assumption restricts the ability of the model to capture complex non-linear anatomical variations. 
In practice, it means that generated shapes are confined to lie within an affine subspace around the mean, which can lead to overly smooth or implausible reconstructions. 
This makes PCAs simple and interpretable, but less practical for representing rich anatomical details. 

\subsection{PointVAE and PointDiffusion: Spatial Deep Generative Models}
While PCA-based statistical shape models provide an interpretable baseline, they are fundamentally limited by their \emph{linearity}. 
Deep generative models, and in particular variational autoencoders (VAEs), offer a natural extension by learning a nonlinear mapping between data and a low-dimensional latent space. 
An encoder $e_\theta$ processes the input signal into a global representation, parameterizing a Gaussian distribution
\[
e_\theta(x) \to \mu, \Sigma, \quad z \sim \mathcal{N}(\mu, \Sigma),
\]
where $z \in \mathbb{R}^d$ denotes the latent code. 
A decoder $d_\gamma$ then maps this latent vector back to the data domain, producing a reconstruction $\hat{x} = d_\gamma(z)$. 
The latent space is regularized to follow a prior distribution, which enables both faithful reconstructions and sampling of new shapes. 
Unlike PCA, the encoder and decoder are modeled by neural networks with non-linear activations, allowing them to capture complex and highly non-linear modes of anatomical variation.  

Spatial deep generative models directly operate on vertex coordinates and connectivity information of 3D shapes. 
Among the various spatial representations explored in the literature, point clouds have emerged as a particularly prominent choice due to their flexibility and efficiency. 
A point cloud representation discards explicit connectivity information and instead models each mesh as an unordered set of points in $\mathbb{R}^3$. 
This eliminates the requirement of consistent mesh connectivity across the dataset and allows a unified treatment of shapes with varying discretizations. 
Here, a 3D mesh is represented by a point cloud $S \subset \mathbb{R}^3$, obtained by uniformly sampling $m$ points from the surface. 
The generative process is modeled using a variational autoencoder (VAE), which maps this set into a latent distribution and reconstructs it back into a set of points approximating the original surface. 
The encoder $e_\theta$ learns to project the input point cloud into the parameters of a Gaussian latent distribution,
\[
e_\theta(S) \to \mu, \Sigma, \quad z \sim \mathcal{N}(\mu, \Sigma),
\]
where $z \in \mathbb{R}^d$ denotes a $d$-dimensional latent vector. The decoder $d_\gamma$ maps the latent code back into an unordered set of points,
\[
\hat{S} = d_\gamma(z) \subset \mathbb{R}^3, \quad |\hat{S}| = m,
\]
aiming to recover the original surface geometry.  

As point clouds are unordered, architectures operating on them must be designed to be invariant to input permutations and extrinsic transformations such as translation, rotation, or scale. 
This invariance is typically enforced by building local neighborhoods through $k$-nearest-neighbor searches or ball queries, and then aggregating features within these neighborhoods using symmetric functions such as max- or mean-pooling. 
PointNet~(\citeauthor{qi:pointnet}) achieves this by applying shared multi-layer perceptrons to each point independently and using a global pooling operator to produce an order-invariant embedding.
Such models scale poorly with input data resolution, which is crucial when analyzing geometries of anatomical shapes.
More recent architectures such as Point Transformers~(\citeauthor{wu:ptv3}) extend this idea by learning attention weights over local neighborhoods defined through spatial proximity, which again requires explicit neighbor queries but allows for richer contextual modeling.
The latest versions of point transformers improve scalability by fast point serialization.
Despite these advances, such networks still operate directly on extrinsic coordinates. 
The geometry of the shape must therefore be inferred from the data, while invariance to global transformations has to be explicitly encoded in the architecture.
The reconstruction loss is defined on unordered sets, most commonly using the Chamfer distance:
\[
d_{\text{CD}}(S, \hat{S}) = 
\sum_{x \in S} \min_{y \in \hat{S}} \|x - y\|_2^2 \; + \;
\sum_{y \in \hat{S}} \min_{x \in S} \|x - y\|_2^2,
\]
which measures the proximity of each point in one set to its nearest neighbor in the other. The full training objective combines this reconstruction loss with the Kullback–Leibler divergence that regularizes the latent space:
\[
\mathcal{L} = d_{\text{CD}}(S, \hat{S}) \; + \; \beta \, \text{KL}\big(\mathcal{N}(\mu, \Sigma) \,\|\, \mathcal{N}(0, I)\big),
\]
with $\beta$ weighting the relative importance of regularization.  

This framework allows for both reconstruction of input shapes and generation of novel samples. 
However, because point clouds are unordered, the network must repeatedly perform expensive neighborhood searches in high-dimensional feature spaces in order to capture local geometric relations. 
When large point clouds are required for anatomical fidelity, the receptive fields must expand proportionally, making training and inference computationally demanding and training is less stable with small datasets at hand. 
Thus, while point-based VAEs provide flexibility, their scalability to high-resolution shapes in low data regimes is limited.

{
Recently, diffusion probabilistic models have emerged as a strong alternative to VAEs for point cloud generation. In particular,}~\cite{luo:point-diffusion} {formulate a point cloud as a set of particles undergoing a gradual forward diffusion process that corrupts data into Gaussian noise, and then learn the corresponding reverse denoising dynamics to generate new shapes.
Unlike VAEs, diffusion models do not require a single-shot mapping from the latent vector to a full point set. Instead, they refine an initial noise point cloud over many steps, which often improves sample fidelity and diversity in point cloud synthesis. The main drawback is computational as sampling requires executing the denoiser for multiple reverse steps. This makes diffusion-based generation substantially slower than a single decoder forward pass, and typically more time demanding in training and inference-time generation.
}

\begin{figure*}[t]
  \centering
  \begin{subfigure}[t]{\textwidth}
    \centering
    \includegraphics[width=0.9\textwidth]{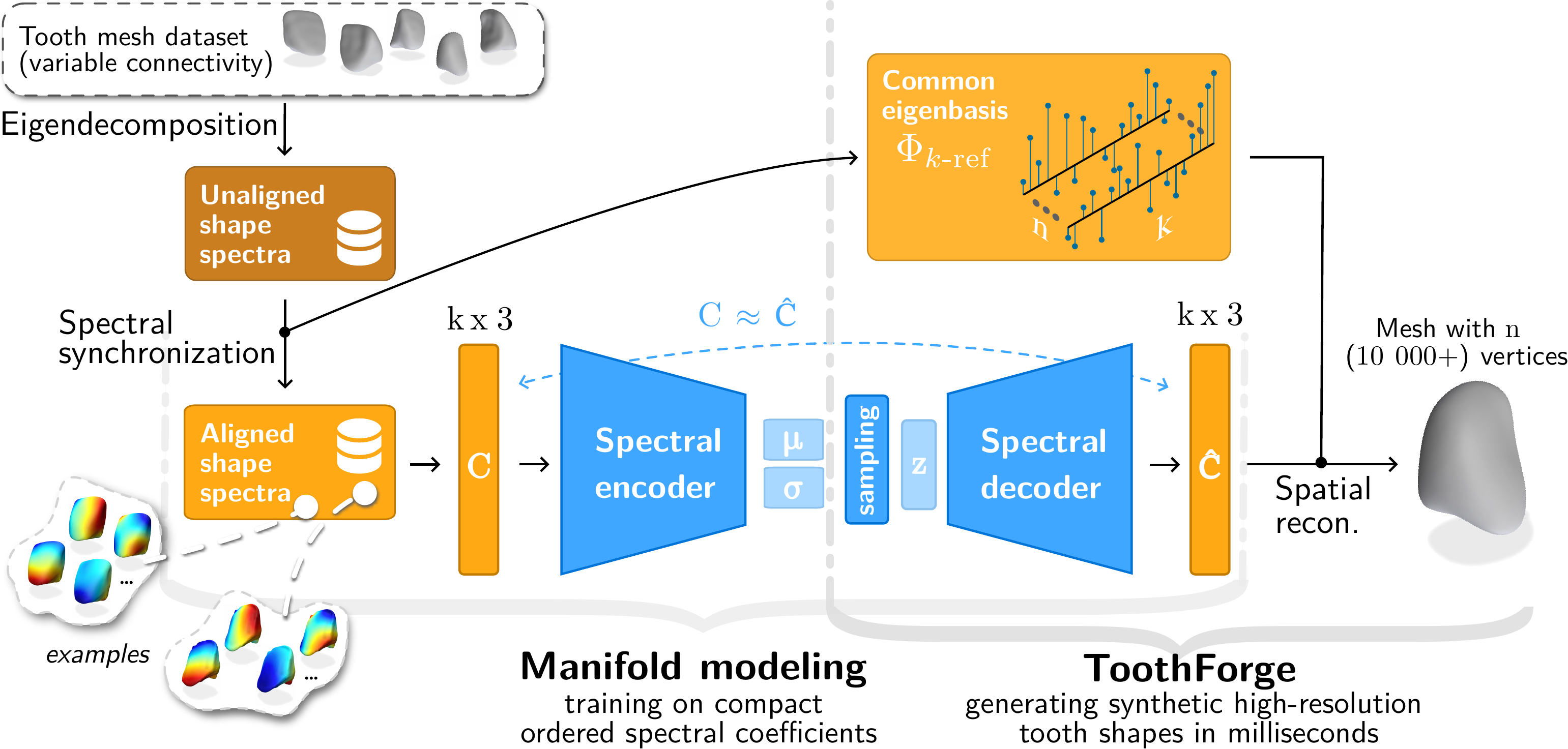}
    \caption{Our framework for generating synthetic shapes of teeth via sampling on a latent manifold. Such manifold is modeled using synchronized spectral coefficients of tooth shapes, denoted as $C$. For
novel data sampling using \emph{ToothForge}, two ingredients are necessary: decoder weights
for infering novel modal coefficient $\hat{C}$ and common eigenbasis $\Phi_{k,\text{ref}}$ to project it to
spatial domain.}
    \label{fig:method-outline}
  \end{subfigure}
  \vspace{1em}
  \begin{subfigure}[t]{\textwidth}
    \centering
    \includegraphics[width=0.96\textwidth]{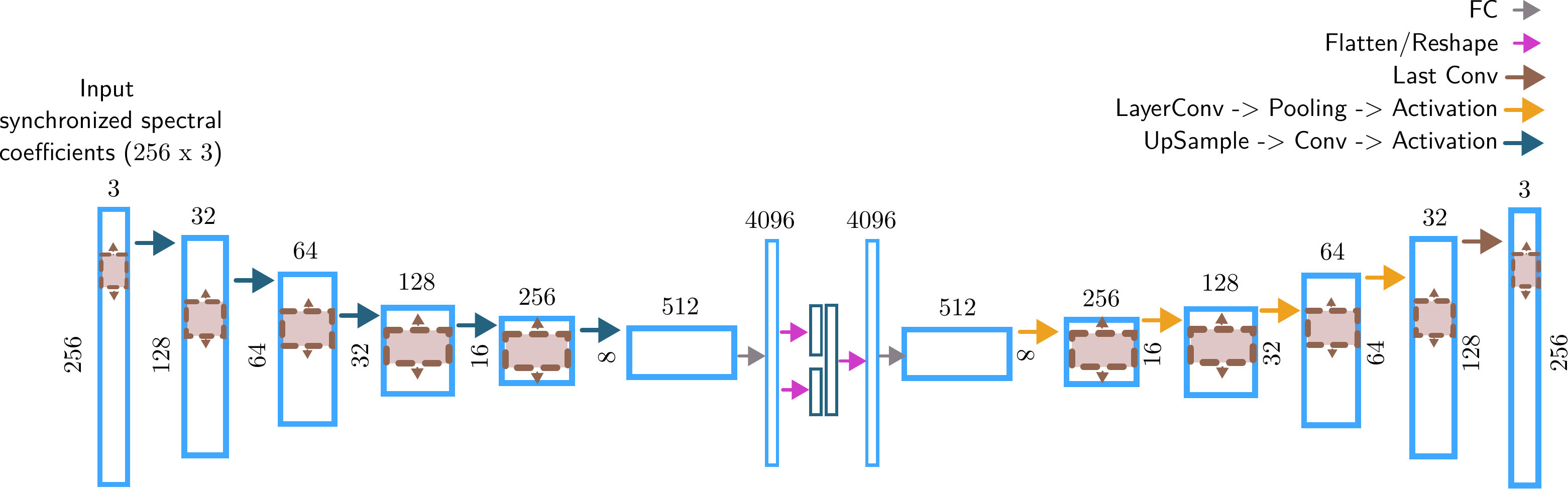}
    \caption{The framework adopts a five-stage 1D convolutional architecture with learnable pooling and upsampling~(\citeauthor{lemeunier:specae}).}
    \label{fig:method-architecture}
  \end{subfigure}
  \caption{%
    Outline and architecture details of ToothForge framework.
  }
  \label{fig:toothforge-method-architecture}
\end{figure*}

\subsection{ToothForge: Spectral Deep Generative Model}
ToothForge is a framework that utilizes spectral coefficients as input features generative deep modeling of dental shape variability.
Instead of operating directly on vertex coordinates, each mesh is projected into the eigenspace of the Laplace–Beltrami operator, providing a compact, intrinsic description of surface geometry. See Figure~\ref{fig:method-outline} for a visual outline. 
Given a closed manifold mesh with vertices $V \in \mathbb{R}^{n \times 3}$, its truncated spectral coefficients are defined as
\[
C_k = \Phi_k^\top V,
\]
where $\Phi_k \in \mathbb{R}^{n \times k}$ contains the first $k$ eigenvectors of the Laplacian, ordered by their associated eigenvalues. 
Lower frequencies capture coarse anatomical structure, such as crown length, while higher frequencies encode fine details, such as molar cusp morphologies, exhibiting rapid oscillations.

A shape can be approximated by reconstruction from these coefficients as
\[
V_k = \Phi_k C_k,
\]
with increasing $k$ yielding higher-fidelity reconstructions.  
This natural ordering of coefficients provides a canonical axis along which neighboring frequencies can be pooled, enabling convolutional and hierarchical operations that are not readily applicable in the unordered spatial domain.

Despite the natural ordering advantage, spectral representations also introduce specific challenges. 
Eigenfunctions are not uniquely defined and may flip signs or swap order across shapes due to instabilities of the decomposition.
ToothForge addresses this by introducing \emph{spectral synchronization}, aligning all spectral embeddings to a common reference basis $\Phi_{k,\text{ref}}$. 
For each shape $M_i$ with coefficients $C_{k,i}$, a transformation $R_i \in \mathbb{R}^{k \times k}$ is estimated such that
\[
\tilde{C}_{k,i} = R_i C_{k,i},
\]
minimizing the discrepancy between the aligned coefficients and the reference basis. 
This procedure eliminates instabilities of the decomposition and provides consistent, connectivity-independent spectral features across the dataset.
The reference template is chosen from the training set and defines the common spectral basis $\Phi_{k,\mathrm{ref}}$ used for synchronization and for projecting decoded coefficients back to the spatial domain. {Importantly, synchronization aligns the eigenbases (i.e., the spectral coordinate system) rather than deforming shapes toward the reference geometry, so the template does not act as a geometric prior. In practice, template can be selected either at random from the training set or as a representative \textit{medoid} shape minimizing average distance to other training samples.}

The aligned spectral coefficients are modeled with a $\beta$-VAE. 
The encoder $e_\theta$ maps coefficients into a Gaussian latent distribution
\[
e_\theta(\tilde{C}_k) \to \mu, \Sigma, \quad z \sim \mathcal{N}(\mu, \Sigma),
\]
and the decoder $d_\gamma$ reconstructs coefficients $\hat{C}_k = d_\gamma(z)$. 
The training objective is a weighted sum of reconstruction error in spectral space and a KL divergence term:
\[
\mathcal{L} = \| \tilde{C}_k - \hat{C}_k \|_2^2 + \beta \, \text{KL}\big(\mathcal{N}(\mu, \Sigma) \,\|\, \mathcal{N}(0, I)\big).
\]  

Although projecting coefficients back into the spatial domain is computationally inexpensive, the optimization is carried out entirely in the spectral domain. 
This reflects a fundamental distinction between \emph{extrinsic} and \emph{intrinsic} learning. 
Spatial generative models must operate on raw coordinates in $\mathbb{R}^3$, and therefore spend capacity accounting for extrinsic factors such as global scale, orientation, translation, and the permutation of input points. 
In contrast, spectral coefficients are intrinsic to the surface and already embed the geometry of the mesh into the representation. 
The network does not need to learn the underlying geometry from data, but can instead focus directly on modeling anatomical variability in a compact and stable space.

\begin{figure}[h!]
    \centering
    \includegraphics[width=\linewidth]{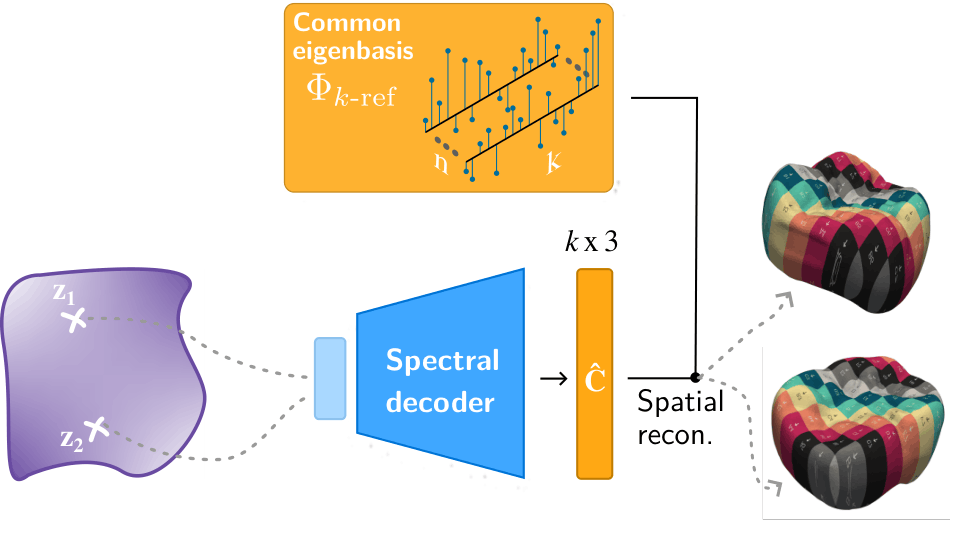}
    \caption{{ToothForge samples new shapes by generating spectral coefficients and reconstructing them through the shared reference eigenbasis. When combined with the template mesh connectivity, all synthesized meshes inherit the same vertex indexing, yielding guaranteed point-to-point correspondence across samples.}}
    \label{fig:consistent-correspondence}
\end{figure}

After training, novel shapes are generated by sampling latent vectors $z \sim \mathcal{N}(0, I)$, decoding them into spectral coefficients $\hat{C}_k$, and projecting back to the spatial domain through the common basis:
\[
\hat{V} = \Phi_{k,\text{ref}} \, \hat{C}_k.
\]
Coupled with the template shape connectivity information, this produces high-resolution meshes with consistent correspondence across samples, {as visualized in Figure}~\ref{fig:consistent-correspondence}. 

ToothForge thereby combines the compactness of spectral embeddings with the generative power of variational autoencoders. 
Compared to point-based VAEs operating on unordered coordinates, ToothForge explicitly leverages intrinsic geometric structure via synchronized spectral coefficients, which is particularly advantageous in bandwidth-matched, low-information regimes.

{We adopt a $\beta$-VAE as the generative backbone because ToothForge is designed to learn a compact, low-dimensional latent manifold and to support efficient sampling for downstream use (e.g., augmentation). Diffusion models are a promising alternative for high-fidelity synthesis, but they typically require iterative sampling and substantially higher training/sampling cost, and are often most effective when trained with larger datasets. We therefore focus on VAEs in this work, and leave diffusion-based generation in the spectral domain as an interesting direction for future research.}

\subsection{Data}
All experiments were performed on a private dataset of 430 dental crown shapes: {149 incisors, 161 premolars and 120 molars.} Data was provided by an industrial partner. 
The dataset was divided into three anatomical categories: incisors, premolars, and molars. 
For each category, an independent model was trained under all evaluated frameworks. 
We train separate models per crown category to maintain anatomically homogeneous shape distributions and to enable a direct, class-wise comparison to standard PCA baselines, which are commonly constructed per anatomical class.

All crowns were digitally modeled by experienced dental technicians with the specific purpose of serving as patient-specific dental prostheses. This ensures clinically realistic anatomical morphology while avoiding biases such as tooth wear or surface irregularities that would be present if shapes were segmented from patient optical scans.

All dental crowns are represented as triangular manifold meshes with various triangulations. 
No texture or color information is included, and the average resolution is $11484 \pm 535$ vertices.
Since the crowns were digitally modeled for use as dental prostheses, the meshes are inherently watertight and free of holes. 
This guarantees that spectral decompositions and correspondence procedures are not affected by artifacts such as holes, self-intersections, or non-manifold edges, ensuring stable downstream processing.

\subsection{Evaluation Metrics}
{All reported geometric distances and reconstruction errors are measured and reported in millimeters (mm).}

\subsubsection{Accuracy}
Accuracy evaluates how well the model can reconstruct shapes from the \emph{training set} after compression. 
When two representations are ordered, such as spatial point clouds with point-to-point correspondence, accuracy is measured using the mean squared error (MSE): 
\[
d_{\text{MSE}}(C_1, C_2) = \| C_1 - C_2 \|_2^2,
\] 
where $C_1, C_2 \subseteq \mathbb{R}^3$ denote the ordered sets.  
For unordered vertex sets, $S_1 \subseteq \mathbb{R}^3$ and $S_2 \subseteq \mathbb{R}^3$, 
we use the Chamfer distance (CD), defined as 
\[
d_{\text{CD}}(S_1, S_2) = 
\sum_{x \in S_1} \min_{y \in S_2} \| x - y \|_2^2 
+ \sum_{y \in S_2} \min_{x \in S_1} \| x - y \|_2^2.
\] 
Lower values of $d_{\text{MSE}}$ and $d_{\text{CD}}$ indicate more faithful reconstructions of the training samples. 
The overall accuracy of a model, denoted $d_{\text{trn-MSE}}$ or $d_{\text{trn-CD}}$, is obtained as the average of the corresponding measure across all shapes in the training set.

\subsubsection{Generalization}
Generalization assesses how well the model can reconstruct \emph{unseen test samples}, thereby measuring its ability to extend beyond the training distribution. 
For point sets with consistent ordering, we compute reconstruction error using the mean squared error (MSE). 
For unordered vertex sets, we instead employ the Chamfer distance (CD), as defined in the previous section. 
Smaller errors correspond to reconstructions that remain faithful to the anatomical details of held-out shapes. 
The reported generalization score of a model, denoted $d_{\text{tst-MSE}}$ or $d_{\text{tst-CD}}$, is obtained by averaging the chosen distance measure across all shapes in the test set.

\subsubsection{Minimum Matching Distance}
While accuracy and generalization quantify how faithfully a model can reconstruct individual shapes after compression, they do not evaluate the quality of the model as a generative distribution. To assess the distributional fidelity of the learned shape space, we evaluate the Minimum Matching Distance (MMD).
MMD evaluates how well the learned generative model captures the fidelity of the training samples using a nearest-neighbor matching score.
Concretely, we draw random latent codes from the prior distribution, decode them into shapes, and obtain a generated set $S_g$ with $|S_g| = |S_r|$, where $S_r$ is the reference set of real shapes. 
We measure the distance of $S_g$ to the reference set $S_r$ of real shapes. 
For each shape in $S_r$, given a set of real crowns $S_r$ and a set of generated crowns $S_g$, we compute the matching error
\[
\text{MMD}(S_r, S_g) = \frac{1}{|S_r|} \sum_{r \in S_r} \min_{g \in S_g} d_{CD}(r, g).
\] 

\noindent Low values indicate that generated samples are on-manifold and do not deviate into unrealistic shapes.

\subsubsection{Coverage}
{To complement MMD, we additionally measure the diversity as the fraction of reference samples that are selected as the nearest neighbor of at least one generated sample:}

\[
\text{COV}(S_r, S_g) = \frac{|\{\text{arg min}_{r \in S_r} d_{CD}(r, g)| g \in S_g\}|}{|S_r|}.
\] 

\noindent {Higher values indicate that the generated set covers a larger portion of the reference set~(\citeauthor{naeem:div-metric-cov}).}

\begin{table*}[t]
\centering
\small
\setlength{\tabcolsep}{5pt}
\caption{Reconstruction and generative quality across spectral truncations $k$ of \emph{ToothForge}.
\emph{SR} (shape recovery) is the average truncation error of the $k$-harmonic reconstruction of the ground-truth meshes for given class and $k$ combination computed between ground-truth vertices $V$ and truncated reconstruction $V_k=\Phi_k\Phi_k^\top V$.
Relative errors ($^\mathrm{rel}$) compare predictions to the $k$-truncated target, absolute errors ($^\mathrm{abs}$) compare to the full-resolution ground truth.
All values are measured with spatial reconstructions using CD as the distance metric.}
\begin{tabular}{l
                ccccc
                ccccc
                ccccc}
\toprule
& \multicolumn{5}{c}{\textbf{Incisors}} 
& \multicolumn{5}{c}{\textbf{Premolars}} 
& \multicolumn{5}{c}{\textbf{Molars}} \\
\cmidrule(lr){2-6}\cmidrule(lr){7-11}\cmidrule(lr){12-16}
$k$ 
& \makebox[1.2cm][c]{SR} 
& $d_{\text{trn}}^{\mathrm{rel}}$ & $d_{\text{trn}}^{\mathrm{abs}}$ 
& $d_{\text{tst}}^{\mathrm{rel}}$ & $d_{\text{tst}}^{\mathrm{abs}}$ 
& \makebox[1.2cm][c]{SR} 
& $d_{\text{trn}}^{\mathrm{rel}}$ & $d_{\text{trn}}^{\mathrm{abs}}$ 
& $d_{\text{tst}}^{\mathrm{rel}}$ & $d_{\text{tst}}^{\mathrm{abs}}$ 
& \makebox[1.2cm][c]{SR} 
& $d_{\text{trn}}^{\mathrm{rel}}$ & $d_{\text{trn}}^{\mathrm{abs}}$ 
& $d_{\text{tst}}^{\mathrm{rel}}$ & $d_{\text{tst}}^{\mathrm{abs}}$ \\
\midrule
32   &  \footnotesize{$0.232$} & \footnotesize{$0.022$} &  \footnotesize{$0.241$} &  \footnotesize{$0.061$} &  \footnotesize{$0.234$}  &  \footnotesize{$0.221$} &  \footnotesize{$0.031$} &  \footnotesize{$0.248$} &  \footnotesize{$0.061$} &  \footnotesize{$0.243$} &  \footnotesize{$0.244$} &  \footnotesize{$0.033$} &  \footnotesize{$0.250$} &  \footnotesize{$0.075$} &  \footnotesize{$0.246$} \\
64   &  \footnotesize{$0.110$} & \footnotesize{$0.041$} &  \footnotesize{$0.152$} &  \footnotesize{$0.066$} &  \footnotesize{$0.155$}   &  \footnotesize{$0.101$} &  \footnotesize{$0.027$} &  \footnotesize{$0.127$} &  \footnotesize{$0.038$} & \footnotesize{$0.129$} &  \footnotesize{$0.123$} & \footnotesize{$0.024$} &  \footnotesize{$0.159$} &  \footnotesize{$0.038$} &  \footnotesize{$0.155$}\\
128  &  \footnotesize{$0.083$} & \footnotesize{$0.046$} &  \footnotesize{$0.123$} &  \footnotesize{$0.060$} &  \footnotesize{$0.128$}  &  \footnotesize{$0.066$} &  \footnotesize{$0.030$} & \footnotesize{$0.079$}  &  \footnotesize{$0.037$} &  \footnotesize{$0.102$} &  \footnotesize{$0.096$} & \footnotesize{$0.033$} &  \footnotesize{$0.093$} &  \footnotesize{$0.033$} &  \footnotesize{$0.108$} \\
256  &  \footnotesize{$0.069$} &   \footnotesize{$0.043$} &  \footnotesize{$0.117$} &  \footnotesize{$0.059$} &  \footnotesize{$0.122$} &  \footnotesize{$0.063$} &  \footnotesize{$0.028$} &  \footnotesize{$0.075$} &  \footnotesize{$0.049$} &  \footnotesize{$0.091$} &  \footnotesize{$0.089$} &  \footnotesize{$0.029$} &  \footnotesize{$0.104$} &  \footnotesize{$0.037$} &  \footnotesize{$0.092$}  \\
512  &  \footnotesize{$0.063$} &  \footnotesize{$0.059$} &  \footnotesize{$0.115$} &  \footnotesize{$0.067$} &  \footnotesize{$0.114$} &  \footnotesize{$0.061$} &  \footnotesize{$0.043$} &  \footnotesize{$0.081$} &  \footnotesize{$0.054$}  &  \footnotesize{$0.102$} &  \footnotesize{$0.084$} &   \footnotesize{$0.030$} &  \footnotesize{$0.099$} &  \footnotesize{$0.043$} &  \footnotesize{$0.093$} \\
1024 &  \footnotesize{$0.062$} & \footnotesize{$0.055$} &  \footnotesize{$0.129$} &  \footnotesize{$0.060$} &  \footnotesize{$0.117$} &  \footnotesize{$0.060$} &  \footnotesize{$0.049$} & \footnotesize{$0.076$} &  \footnotesize{$0.063$} &  \footnotesize{$0.100$} &  \footnotesize{$0.084$} &   \footnotesize{$0.033$} &  \footnotesize{$0.094$} &  \footnotesize{$0.071$} &  \footnotesize{$0.101$}  \\
\bottomrule
\end{tabular}
\label{tab:toothforge-results}
\end{table*}

\section{Results}
\subsection{Experimental Setup}
Dense correspondence was first established among all meshes to enable construction of the PCA-based statistical shape model.
A single template was selected, simplified to $4096$ vertices, and normalized to unit scale. For each remaining mesh, $4096$ surface points were sampled and aligned to the template using rigid ICP~\cite{rusinkiewicz:icp}, followed by non-rigid coherent point drift from~\cite{myronenko:cpd} with parameters $\beta=2.0$, $\lambda=3.0$, and 150 iterations. 
This produced a set of meshes sharing the same template connectivity. 
The registered meshes were vectorized by concatenating their $x,y,z$ coordinates and stacked into a shape matrix. 
Shapes were centered and normalized before PCA was applied with $k$ retained components. 
Novel shapes were generated by sampling coefficients $b_i \sim \mathcal{N}(0, \lambda_i)$ and forming linear combinations of principal components with the mean shape.

As nonlinear spatial baselines, a point-based variational autoencoder {(PointVAE) was used, similar to what is presented in}~\cite{becker:point-deep-models}. {The encoder and decoder was replaced by PointNet++}~(\citeauthor{qi:pnetpp}) and implemented with architecture matched in depth to ToothForge.
The network follows a 5-stage encoder-decoder design with a Gaussian bottleneck of dimension $d=16$, optimized using Chamfer distance on vertex positions combined with a $\beta$-weighted KL divergence term. 
A cyclical annealing of $\beta$ between 0 and 0.05 was applied. 
Input point clouds were sampled using Poisson disk sampling with $m$ points per mesh. 
Training was performed with batch size $16$ using AdamW (initial learning rate $10^{-4}$, cosine annealing restarts every $10000$ iterations) for approximately two hours on a single Tesla T4 GPU. 
{Second point-based framework is a point diffusion model presented by}~\citeauthor{luo:point-diffusion}. {More specifically, we employ their point cloud generator model with normalizing flows. We keep most of the hyperparameters as proposed defaults, and only change the latent dimensionality to $64$. Input point clouds were sampled using Poisson disk sampling with $m$ points per mesh. We optimized the model for approximately two hours using single Tesla T4 GPU.
We refer to this model as PointDiffusion in the text.}
To encourage robustness in both spatial approaches, small random perturbations in global rotation and isotropic scale were applied to the input point clouds during training.
Separate models were trained for incisors, premolars, and molars, using the same 80/20 train–test split as with PCA.

ToothForge operates in the spectral domain, where each mesh is projected into the $k$-truncated eigenbasis of the Laplace–Beltrami operator. 
Spectral embeddings are synchronized to a common reference basis using the method of~\cite{lombaert:brain-transfer}, ensuring stable and consistent coefficients across shapes. 
A $\beta$-VAE with a 5-stage 1D convolutional encoder–decoder, latent size $d=16$, and cyclical annealing of $\beta$ between 0 and 0.05 was trained.
The architecture uses pooling and unpooling operators as presented by~\cite{lemeunier:specae}.
See Figure~\ref{fig:method-architecture} for details in scenario where $256$ modes are used during training.
The reconstruction loss was defined purely in the spectral domain, combining mean squared error on coefficients with the KL divergence term. Training was performed with batch size $16$ using AdamW (initial learning rate $10^{-4}$, cosine annealing restarts every 10,000 iterations) for approximately two hours on a Tesla T4 GPU. 
Separate models were trained for incisors, premolars, and molars using the same 80/20 train–test split as for the other baselines.

\subsection{Results on ToothForge}
\subsubsection{Reconstruction Quality Across Tooth Classes and Spectral Truncations}
A central design choice in spectral generative modeling is the number of Laplace–Beltrami eigenfunctions retained when projecting a mesh into the spectral domain. 
Lower values of $k$ yield a more compact representation but may discard high-frequency information, while larger values preserve fine geometric detail at the cost of higher dimensionality. 
In practice, $k$ is selected based on the target trade-off between compactness and geometric detail, e.g., by choosing the smallest $k$ for which reconstruction/coverage metrics (and the corresponding qualitative fidelity) have saturated for the intended application.
To assess how this trade-off influences ToothForge, reconstruction quality was quantified across different truncation levels.  

Reconstruction errors are reported in two complementary ways. 
First, predictions are compared against the $k$-truncated meshes used to generate the training coefficients. 
This relative error isolates the fidelity of the network. 
Second, reconstructions are compared against the original high-resolution ground-truth meshes.
This absolute error captures both truncation and network reconstruction losses, reflecting the overall clinical fidelity of the model.  
The results are summarized in Table~\ref{tab:toothforge-results}.

\begin{figure*}[t!]
    \centering
    \includegraphics[width=0.82\linewidth]{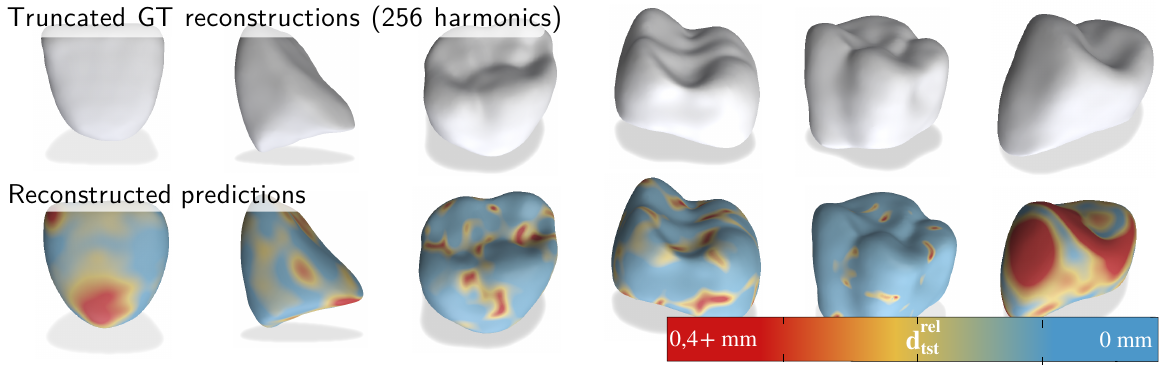}
    \caption{{Reconstructions of unseen tooth shapes. The predictions accurately capture the overall tooth shape. The reconstructions may appear smoothed at times (rightmost premolar). This is due to possible inaccuracies in predicting high-frequency components.}}
    \label{fig:recons-w-error-maps}
\end{figure*}

The relative reconstruction errors indicate that the model generalizes well to unseen samples across tooth classes. 
As the number of spectral harmonics increases, the reconstruction quality remains stable, with only a slight rise in the relative training error. 
This trend suggests that while the model effectively captures the global geometric structure, it becomes increasingly sensitive to the higher-frequency spectral components. 
These components are inherently less stable and more susceptible to noise. 
Consequently, the spectral representation of fine geometric details may exhibit minor irregularities, which can challenge the network’s ability to model them consistently.
The absolute reconstruction errors remain consistently low across all tooth classes, reflecting high fidelity to the full-resolution ground truth. 
Their gradual reduction with increasing $k$ indicates that additional spectral components improve geometric detail recovery without introducing instability in the spatial domain.
Although minor noise appears at higher $k$ values, the absolute errors continue to approach the $SR$ values.
This indicates that most of the remaining discrepancy originates from the intrinsic low-pass filtering rather than from network reconstruction errors. {See Figure}~\ref{fig:recons-w-error-maps} {for randomly selected tooth reconstructions.}

\subsubsection{Ablation Study}
A quantitative ablation study quantified the effect of individual design choices, with results summarized in Table~\ref{tab:ablation}. 
The parameter $k$ was fixed at $256$, and analysis focused on the molar class, noting that similar trends hold across other truncation values and tooth classes.
Setting $\beta = 0$ eliminated latent regularization and led to reconstructions that collapsed toward overly smooth mean-shape-like geometries, with little variability across latent space exploration. 
Removing spectral synchronization consistently degraded reconstructions: in the spatial domain, generated outputs were often semantically far from dental structures, producing implausible shapes.
As an additional test, the training loss was modified to include a spatial regularization term, implemented as an auxiliary Chamfer distance on reconstructed vertices,
\[
\mathcal{L}
= \bigl\| \tilde{C}_k - \hat{C}_k \bigr\|_2^2
+ \beta\,\mathrm{KL}\!\Bigl(\mathcal{N}(\mu,\Sigma)\,\big\|\,\mathcal{N}(0,I)\Bigr)
+ \lambda_{\mathrm{sp}}\, d_{\mathrm{CD}}(V,\hat{V}),
\]
where $V$ and $\hat{V}$ denote the ground-truth and reconstructed vertex sets, respectively, and $d_{\mathrm{CD}}(\cdot,\cdot)$ is the Chamfer distance as defined earlier. 
Across all tooth classes, this additional spatial term did not yield consistent improvements and in several cases slightly worsened results. 
Overall differences were neglectable. 
These findings support that the intrinsic smoothness encoded by the spectral representation is sufficient, and that adding explicit spatial regularization does not benefit ToothForge.

\begin{table}[t]
\centering
\small
\caption{Ablation at $k{=}256$ for molar class. All values are measured with spatial reconstructions using CD as the distance metric. \textbf{Bold} values denote the best-performing results and \underline{underlined} values indicate the second-best within each column.}
\begin{tabular}{l@{\hspace{3.5em}}cccc}
\toprule
Variant & $d_{\text{trn}}^{\mathrm{rel}}$ & $d_{\text{tst}}^{\mathrm{rel}}$ & $d_{\text{tst}}^{\mathrm{abs}}$ & $MMD$ \\
\midrule
ToothForge      & \footnotesize{$0.029$} & \footnotesize{$\textbf{0.037}$} & \footnotesize{$\textbf{0.092}$} & \underline{\footnotesize{$0.099$}} \\
$\beta = 0$                   & \footnotesize{$0.066$} & \footnotesize{$0.125$} & \footnotesize{$0.232$} & \footnotesize{$0.212$} \\
No spec. sync.   & \footnotesize{$0.101$} & \footnotesize{$0.127$} & \footnotesize{$0.160$} & \footnotesize{$0.186$} \\
+$\lambda_{\text{sp}}{=}1\!\times\!10^{-2}$ & \footnotesize{$0.027$} & \footnotesize{$0.041$} & \underline{\footnotesize{$0.105$}} & \footnotesize{$0.111$} \\
+$\lambda_{\text{sp}}{=}5\!\times\!10^{-2}$ & \footnotesize{$\textbf{0.022}$} & \underline{\footnotesize{$0.040$}} & \footnotesize{$0.112$} & \footnotesize{$0.102$} \\
+$\lambda_{\text{sp}}{=}1\!\times\!10^{-3}$ & \underline{\footnotesize{$0.026$}} & \footnotesize{$0.044$} & \footnotesize{$0.108$} & \footnotesize{$\textbf{0.098}$} \\
\bottomrule
\end{tabular}
\label{tab:ablation}
\end{table}

\begin{figure*}[t!]
    \centering
    \includegraphics[width=\linewidth]{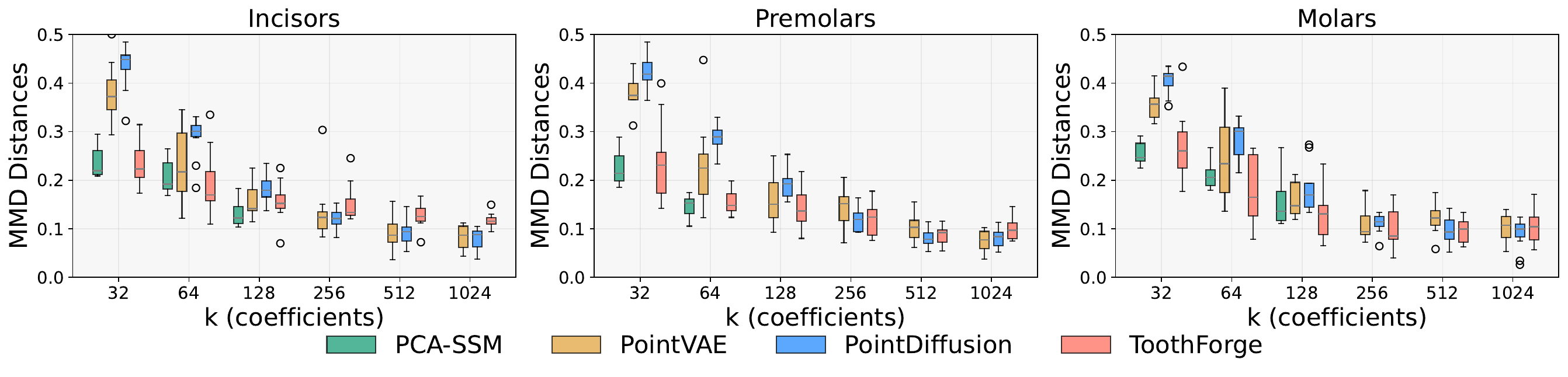}
    \caption{Reconstruction fidelity across spectral truncations $k$ for molars using PCA, PointVAE, PointDiffusion and ToothForge. Note that PCA results are not reported for higher $k$ values due to the insufficient number of available samples to support the corresponding number of principal components. For the PointVAE and PointDiffusion, $k$ denotes the number of input spatial points ($x$, $y$, $z$).}
    \label{fig:comparative-quantitative}
\end{figure*}

\subsubsection{Robustness to Template Shape Selection}
\begin{figure}[h!]
    \centering
    \includegraphics[width=\linewidth]{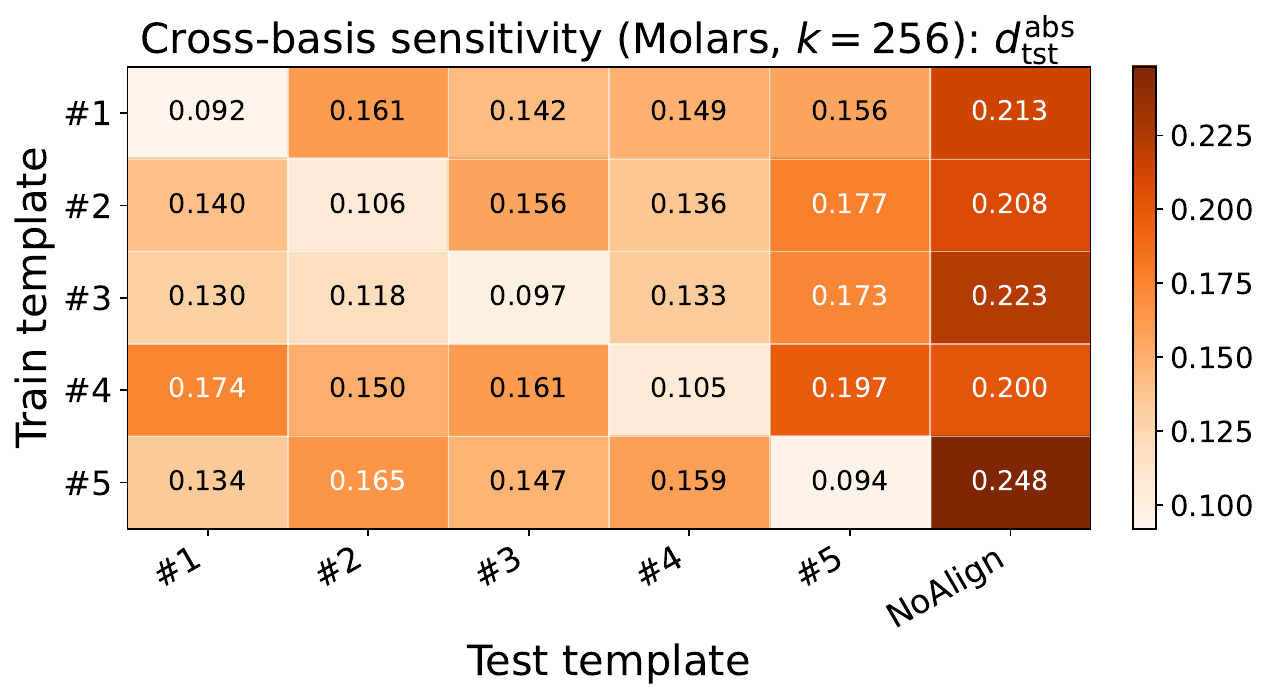}
    \caption{Cross-basis sensitivity for the molar class at $k = 256$. Heatmap entries show the absolute test reconstruction error $d_{\text{tst}}^{\mathrm{abs}}$ when training with one reference template (rows) and evaluating test shapes aligned to another template (columns). Diagonal entries correspond to matched train–test templates and yield the lowest errors. Off-diagonal entries show modest degradation when train and test templates differ. The “NoAlign” column indicates performance without spectral synchronization, resulting in substantially higher errors.}
    \label{fig:template-robustness}
\end{figure}

The robustness of ToothForge to the choice of reference template was evaluated by training models on different randomly selected template shapes and testing them across all possible train–test template combinations. 
The results are presented in Figure~\ref{fig:template-robustness}.
The diagonal entries, where training and test data were synchronized to the same template, consistently produced the lowest reconstruction errors and plausible reconstructions.
In contrast, the off-diagonal cases, where different templates were used for training and testing, showed that the generated shapes were less accurate, underlining that mismatched bases introduce variability into the predictions.
While this degradation was not as severe as in the “NoAlign” setting, the results indicate that stable performance is primarily achieved when train and test shapes are synchronized to the same reference template.
We observed similar trends across tooth classes and $k$ values.
Importantly, this experiment indicates that the reference template does not bias the generated geometry toward a particular shape. Instead, it establishes a common spectral basis used to express coefficients consistently. The main failure mode arises from using different bases between training and inference, rather than from any particular choice of template. In practice, reference can be chosen from the training set as a representative shape (e.g., a medoid) or at random, and then kept fixed throughout training and generation to ensure a stable synchronization convention.

\subsection{Comparative Analysis}
\begin{figure*}[t!]
    \centering
    \includegraphics[width=\linewidth]{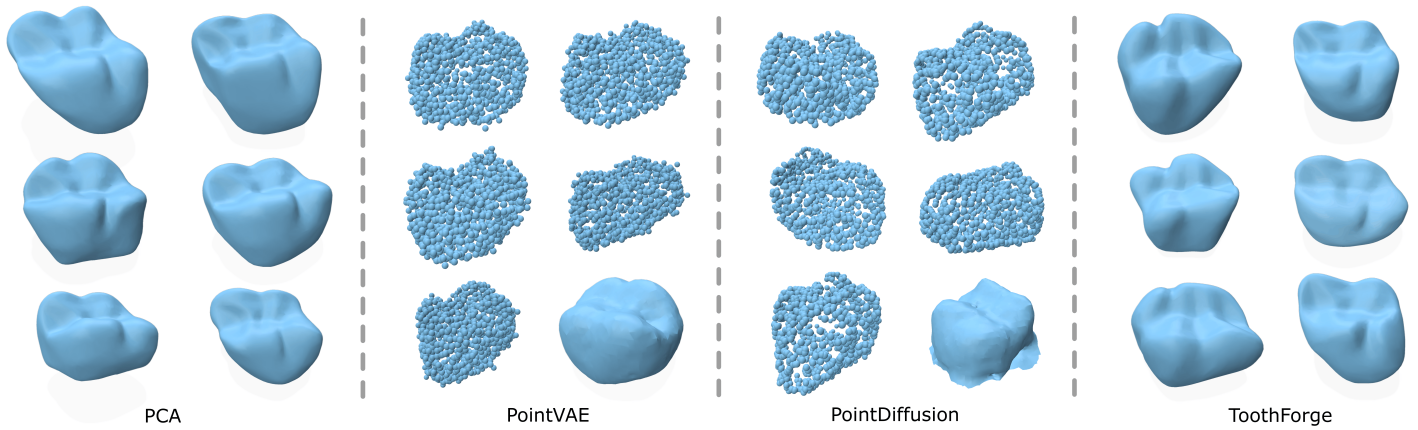}
    \caption{{Samples from the evaluated methods. For PCA and ToothForge, we visualize reconstructed meshes directly using the shared template connectivity and per-vertex correspondence. For point-based methods, we display raw point clouds, with one representative sample additionally reconstructed via Poisson surface reconstruction. Note that while both PCA and ToothForge capture global crown characteristics well, PCA-generated samples exhibit reduced diversity in localized anatomical features like lingual/buccal grooves, whereas ToothForge yields a broader range of fine-scale morphological variations. Point-based samples tend to show more pronounced local surface noise.}}
    \label{fig:comparative-qualitative}
\end{figure*}

First, quantitative comparisons are provided in Figure~\ref{fig:comparative-quantitative} and Table~\ref{tab:comparative-quantitative}.
ToothForge consistently achieves low minimal matching distance values on molar class. Same trend was observed on incisors and premolars.
{For experiments with low truncations ($k = 32, 64$), the limited number of coefficients constrains the representation capacity, leading to reconstructions that are coarse and not anatomically plausible. PointVAE and PointDiffsion show higher error in low $k$ setup because its inputs are sparse point clouds (matched in bandwidth to the coefficient dimensionality), providing substantially less geometric information for reconstruction than the spectral representation.
For higher $k$ values, both spectral and point-based methods capture tooth shape variability well. 
PointVAE, however, often produces locally noisy surfaces. 
PointDiff exhibits strong generative behavior, but sampling is substantially slower (milliseconds for a single forward pass of ToothForge vs. seconds for iterative diffusion), which limits practical throughput.
Moreover, these point-based frameworks produce unordered point sets without connectivity, so a separate surface reconstruction step (e.g., Poisson reconstruction) is required. When applied to sparse point clouds, this post-processing can yield implausible or anatomically distorted meshes.}
In contrast, eventhough spectral reconstructions tend to be slightly smoothed, they directly yield coherent surfaces, which better preserves anatomical fidelity and results in more plausible crown geometries.
The difference in reconstruction quality is visualized in Figure~\ref{fig:comparative-qualitative}.
As for the comparison with PCA, both PCA and ToothForge capture global crown characteristics well. However, PCA-generated samples show reduced diversity in localized anatomical features, most notably in fine-scale regions such as lingual/buccal grooves, whereas ToothForge yields a broader range of such morphological variations.

Jointly, the quantitative and qualitative results show at least two things.
{First, in the context of population-level tooth modeling, the benefits of non-linear latent spaces are not markedly pronounced compared to classical statistical shape models such as PCA.
Nevertheless, our findings demonstrate that non-linear spectral modeling can achieve comparable reconstruction fidelity while leveraging a compact, frequency-ordered intrinsic coefficient representation that is stable under changes in discretization through synchronization.
ToothForge concentrates learning on a fixed-length intrinsic descriptor whose dimensionality is decoupled from mesh resolution, while remaining compatible with future extensions such as conditional generative modeling}~(\citeauthor{zhang:conddiff}).
{This could enable more practical frameworks, for example by learning a single latent space for all tooth classes or by introducing attribute-guided shape synthesis, text-driven generating}~(\citeauthor{xu:pointllm}), {or geometry modifying}~(\citeauthor{achlioptas:shapetalk}).
We also show that the distributional coverage and fidelity is on par with point-based frameworks, but without its limitations in representing fine anatomical details in low-data settings. 
While even more advanced point-based generative methods for dental structures~(\citeauthor{chanintonsongkhla:2025latent}) can capture geometric complexity through point-flow mechanisms~(\citeauthor{yang:pointflow}), they require extensive optimization (30 days for anterior teeth) and large datasets. 
This compactness is therefore a key distinction from spatial generative models, which typically operate directly on high-dimensional, unordered coordinates and therefore require substantially more model capacity, computational time, and bigger datasets to learn invariances and local structure.

\begin{table*}[t]
\centering
\small
\caption{{Quantitative comparison and runtime of the evaluated generative methods on the molar tooth class. We report Minimum Matching Distance (MMD, $\downarrow$) and Coverage (COV, $\uparrow$) computed using Chamfer Distance (CD), together with training and sampling time. PCA uses the top 128 input features, while deep learning baselines use 256 features. Best results are highlighted in \textbf{bold}, and second-best results are} \underline{underlined}.}
\begin{tabular}{l@{\hspace{3.5em}}cccc}
\toprule
Method & MMD & COV (\%) & Training Time & Sampling Time \\
\midrule
PCA         & \footnotesize{$0.1541$}               & \footnotesize{$37.04$}                & \footnotesize{\textbf{$\sim\textbf{10}$\,s}}  & \underline{\footnotesize{$\sim10$\,ms}} \\
PointVAE    & \underline{\footnotesize{$0.1090$}}   & \footnotesize{$42.63$}                & \footnotesize{$\sim121$\,mins}                & \footnotesize{$\sim22$\,ms}\\
PointDiffusion  & \footnotesize{$0.1112$}               & \footnotesize{$\textbf{47.00}$}       & \footnotesize{$\sim143$\,mins}    & \footnotesize{$\sim9$\,s}\\
ToothForge  & \footnotesize{$\textbf{0.0997}$}      & \underline{\footnotesize{$43.78$}}    & \underline{\footnotesize{$\sim100$\,mins}}                & \footnotesize{\textbf{$\sim\textbf{1}$\,ms}}\\
\bottomrule
\end{tabular}
\label{tab:comparative-quantitative}
\end{table*}

\subsection{Downstream Task Experiment}
{To assess whether ToothForge is useful beyond shape reconstruction, we evaluate it as a data augmentor in a tooth class classification (incisor/premolar/molar). Because this task is relatively easy on the full dataset, we deliberately restrict the classifier training set to $100$ randomly selected shapes while preserving the original class distribution, making the setting data-limited and thus more sensitive to augmentation.
To prevent any data contamination, ToothForge is trained only on the remaining training split, excluding the $100$ shapes used to train the classifier. The test set is identical for generative and classification task.

We compare setups with no augmentation, conventional mesh-space augmentations only, ToothForge samples only, and both combined. For conventional augmentations, we apply a rigid transform, random rotation around the z-axis up to $45\degree$ and, a deformation augmentation, vertex jittering with Gaussian noise ($\sigma = 0.005$). For ToothForge augmentation, wer consider two budgets $100$ and $1000$ generated shapes added to the training set.

Table}~\ref{tab:downstream-task} {shows that ToothForge can serve as an effective augmentor in this low-data setting. Training without augmentation yields $90.36\%$ accuracy, while conventional mesh-space augmentations improve this to $95.18\%$. Using $1000$ ToothForge samples leads to a larger gain, reaching $98.79\%$, and adding mesh transformations on top does not further change performance, indicating near-ceiling accuracy on this simple task and relatively small test set.

To better probe complementarity before saturation, we also evaluate a smaller generative budget. With only $100$ ToothForge samples, accuracy increases to $92.77\%$, i.e., better than no augmentation but below conventional augmentations. Importantly, combining these $100$ generated samples with mesh transformations improves accuracy to $96.38\%$, outperforming mesh transformations alone ($95.18\%$). This suggests that ToothForge provides additional, non-trivial shape variability that complements standard rigid and deformation augmentations.}

\begin{table}[t]
\centering
\small
\caption{{Downstream tooth type classification (incisor/premolar/molar) under different augmentation strategies in the low-data regime (100 training shapes, class-balanced). Mesh transforms denotes random z-axis rotations and vertex jitter. ToothForge augmentation adds 100 or 1000 generated crowns sampled from the learned manifold. Accuracy is reported on a fixed held-out test set. Best results are highlighted in \textbf{bold}, and second-best results are }\underline{underlined}.}
\begin{tabular}{lc}
\toprule
Variant & Accuracy (\%) \\
\midrule
No augmentation                                 & \footnotesize{$90.36$} \\
Mesh transforms (rotation + vertex jitter)      & \underline{\footnotesize{$95.18$}} \\
ToothForge ($N=1000$)                           & \footnotesize{$\textbf{98.79}$} \\
Mesh transforms + ToothForge ($N=1000$)         & \footnotesize{$\textbf{98.79}$} \\

\midrule

Mesh transforms (rotation + vertex jitter)      & \underline{\footnotesize{$95.18$}} \\
ToothForge ($N=100$)                            & \footnotesize{$92.77$} \\
Mesh transforms + ToothForge ($N=100$)          & \footnotesize{$\textbf{96.38}$} \\

\bottomrule
\end{tabular}
\label{tab:downstream-task}
\end{table}

\section{Conclusion}
This study presented ToothForge, a deep generative framework that learns dental crown geometry from synchronized spectral representations.
By operating in the spectral domain, the method models intrinsic shape variability through compact and interpretable latent spaces that remain stable across heterogeneous mesh connectivities.
Comprehensive experiments across incisor, premolar, and molar classes demonstrated consistent reconstruction accuracy, valided across varying number of spectral modes used during training.
Our findings indicate that synchronized spectral modeling exhibits robustness to reference selection, as long as a consistent synchronization framework is maintained between training and testing.
Compared to PCA and point-based generative models, ToothForge provides a good balance of compactness, fast sampling, and strong generative performance.
Finally, we showed that ToothForge is effective as a generative data augmentor, improving performance in a low-data downstream tooth-type classification task and complementing standard mesh-space augmentations.
Collectively, these findings validate the efficacy of spectral embedding methodologies for medical shape generation tasks, particularly under the constraints of limited data availability and high-resolution representation characteristic of digital dentistry applications.
{We emphasize that the main value of ToothForge is not solely improved reconstruction over PCAs under ideal correspondence, but a practical generative representation for clinical datasets where consistent connectivity and discretization cannot be guaranteed. By combining spectral synchronization with compact, frequency-ordered coefficients, ToothForge preserves PCA-like efficiency while enabling non-linear generative modeling in a space that is readily extensible. 
This creates a direct path toward (i) unified latent space that can model multiple classes like incisors, premolars, and molars jointly, (ii) conditional generation where we can steer synthesis using explicit attributes like target crown size or cusp prominence), and (iii) controllable shape editing workflows that modify an existing crown using additional inputs such as partial geometry, occlusal constraints, or other multimodal clinical signals.}

\ethics{The work follows appropriate ethical standards in conducting research and writing the manuscript, following all applicable laws and regulations regarding treatment of animals or human subjects.}

\coi{We declare we don't have conflicts of interest.}

\data{The data used in this study consist of patient-specific dental crown prostheses provided by an industrial partner. Due to privacy regulations and data protection agreements, these data cannot be publicly shared.}

\bibliography{sample}

@INPROCEEDINGS{rusinkiewicz:icp,
  author={Rusinkiewicz, S. and Levoy, M.},
  booktitle={International Conference on 3D Digital Imaging and Modeling}, 
  title={Efficient variants of the ICP algorithm}, 
  year={2001}
}

@article{myronenko:cpd,
author = {Myronenko, Andriy and Song, Xubo},
year = {2010},
title = {Point Set Registration: Coherent Point Drift},
journal = {IEEE Transactions on Pattern Analysis and Machine Intelligence (PAMI)},
}

@InProceedings{kubik:toothforge,
author="Kub{\'i}k, Tibor
and Guibault, Fran{\c{c}}ois
and {\v{S}}pan{\v{e}}l, Michal
and Lombaert, Herv{\'e}",
title="ToothForge: Automatic Dental Shape Generation Using Synchronized Spectral Embeddings",
booktitle="Information Processing in Medical Imaging (IPMI)",
year="2026"
}

@InProceedings{lombaert:brain-transfer,
author="Lombaert, Herve
and Arcaro, Michael
and Ayache, Nicholas",
title="Brain Transfer: Spectral Analysis of Cortical Surfaces and Functional Maps",
booktitle="Information Processing in Medical Imaging (IPMI)",
year="2015",
}

@inproceedings{qi:pnetpp,
author = {Qi, Charles R. and Yi, Li and Su, Hao and Guibas, Leonidas J.},
title = {PointNet++: deep hierarchical feature learning on point sets in a metric space},
year = {2017},
booktitle = {Neural Information Processing Systems (NeurIPS)},
}

@inproceedings{zhang:conddiff,
    author    = {Zhang, Lvmin and Rao, Anyi and Agrawala, Maneesh},
    title     = {Adding Conditional Control to Text-to-Image Diffusion Models},
    booktitle = {International Conference on Computer Vision (ICCV)},
    year      = {2023},
}

@article{chanintonsongkhla:2025latent,
  title={A latent variable deep generative model for 3D anterior tooth shape},
  author={Chanintonsongkhla, Chawalit and Chouvatut, Varin and Bunkhumpornpat, Chumphol and Theerasopon, Pornpat},
  journal={Journal of Prosthodontics},
  year={2025}
}

@inproceedings{yang:pointflow,
  title={Pointflow: 3d point cloud generation with continuous normalizing flows},
  author={Yang, Guandao and Huang, Xun and Hao, Zekun and Liu, Ming-Yu and Belongie, Serge and Hariharan, Bharath},
  booktitle={International Conference on Computer Vision (ICCV)},
  year={2019}
}

@article{nauwelaers:palatal-variation-dl,
author = {Nauwelaers, Nele and Matthews, Harold and Fan, Yi and Croquet, Balder and Hoskens, Hanne and Mahdi, Soha and El Sergani, Ahmed and Gong, Shunwang and Xu, Tianmin and Bronstein, Michael and Marazita, Mary and Weinberg, Seth and Claes, Peter},
year = {2021},
title = {Exploring palatal and dental shape variation with 3D shape analysis and geometric deep learning},
journal = {Orthodontics \& Craniofacial Research},
}

@article{primovzivc:palatal-clinical,
  title={Early crossbite correction: a three-dimensional evaluation},
  author={Primo{\v{z}}i{\v{c}}, Jasmina and Ovsenik, Maja and Richmond, Stephen and Kau, Chung How and Zhurov, Alexei},
  journal={European Journal of Orthodontics},
  year={2009},
}

@article{matsuyama:palatal-clinical-ii,
  title={Effects of palate depth, modified arm shape, and anchor screw on rapid maxillary expansion: a finite element analysis},
  author={Matsuyama, Yosuke and Motoyoshi, Mitsuru and Tsurumachi, Niina and Shimizu, Noriyoshi},
  journal={European Journal of Orthodontics},
  year={2015},
}

@article{cleghorn:endo-clinical-i,
  title={Root and root canal morphology of the human permanent maxillary first molar: a literature review},
  author={Cleghorn, Blaine M and Christie, William H and Dong, Cecilia CS},
  journal={Journal of endodontics},
  year={2006},
}

@article{peiris:endo-clinical-ii,
  title={Root canal morphology of mandibular permanent molars at different ages},
  author={Peiris, HRD and Pitakotuwage, TN and Takahashi, M and Sasaki, K and Kanazawa, E},
  journal={International Endodontic Journal},
  year={2008},
}

@article{bhola:implants-clinical-i,
  title={Immediate implant placement: clinical decisions, advantages, and disadvantages},
  author={Bhola, Monish and Neely, Anthony L and Kolhatkar, Shilpa},
  journal={Journal of Prosthodontics: Implant, Esthetic and Reconstructive Dentistry},
  year={2008},
}

@article{testori:implants-clinical-ii,
  title={Implant placement in the esthetic area: criteria for positioning single and multiple implants},
  author={Testori, Tiziano and Weinstein, Tommaso and Scutell{\`a}, Fabio and Wang, Hom-Lay and Zucchelli, Giovanni},
  journal={Periodontology},
  year={2018},
}

@article{kong:ai-prosthesis,
  title={Application of artificial intelligence in dental crown prosthesis: a scoping review},
  author={Kong, Hyun-Jun and Kim, Yu-Lee},
  journal={BMC Oral Health},
  year={2024},
}

@InProceedings{golriz:crown-generation,
    author      = {Hosseinimanesh, Golriz and 
                   Ghadiri, Farnoosh and 
                   Guibault, Francois and 
                   Cheriet, Farida and 
                   Keren, Julia},
    title       = {From Mesh Completion to AI Designed Crown},
    booktitle   = {Medical Image Computing and Computer Assisted Intervention (MICCAI)},
    year        = {2023},
}

@InProceedings{yang:crown-generation,
    author      = {Yang, Su and 
                   Han, Jiyong and 
                   Lim, Sang-Heon and 
                   Yoo, Ji-Yong and 
                   Kim, SuJeong and 
                   Song, Dahyun and 
                   Kim, Sunjung and 
                   Kim, Jun-Min and 
                   Yi, Won-Jin},
    title        = {DCrownFormer: Morphology-Aware Point-to-Mesh Generation Transformer for Dental Crown Prosthesis from 3D Scan Data of Antagonist and Preparation Teeth},
    booktitle    = {Medical Image Computing and Computer Assisted Intervention (MICCAI)},
    year         = {2024}
}

@Article{tian:crown-generation,
    author       = {Tian, Sukun and 
                    Huang, Renkai and
                    Li, Zhenyang and 
                    Fiorenza, Luca and 
                    Dai, Ning and 
                    Sun, Yuchun and 
                    Ma, Haifeng},
    title        = {A dual discriminator adversarial learning approach for dental occlusal surface reconstruction},
    journal      = {Journal of Healthcare Engineering},
    year         = {2022},
}

@article{wold:pca,
title = {Principal component analysis},
journal = {Chemometrics and Intelligent Laboratory Systems},
year = {1987},
author = {Svante Wold and Kim Esbensen and Paul Geladi},
}

@inproceedings{pan:variational-point-ae,
  title={Variational relational point completion network},
  author={Pan, Liang and Chen, Xinyi and Cai, Zhongang and Zhang, Junzhe and Zhao, Haiyu and Yi, Shuai and Liu, Ziwei},
  booktitle={Conference on Computer Vision and Pattern Recognition (CVPR)},
  year={2021}
}

@InProceedings{adams:pcssms,
author="Adams, Jadie
and Elhabian, Shireen Y.",
title="Can Point Cloud Networks Learn Statistical Shape Models of Anatomies?",
booktitle="Medical Image Computing and Computer Assisted Intervention (MICCAI)",
year="2023",
}

@inproceedings{yang:foldingnet,
  title={Foldingnet: Point cloud auto-encoder via deep grid deformation},
  author={Yang, Yaoqing and Feng, Chen and Shen, Yiru and Tian, Dong},
  booktitle={Conference on Computer Vision and Pattern Recognition (CVPR)},
  year={2018}
}

@article{chen:vae-disentanglement,
  title={Isolating sources of disentanglement in variational autoencoders},
  author={Chen, Ricky TQ and Li, Xuechen and Grosse, Roger B and Duvenaud, David K},
  journal={Neural Information Processing Systems (NeurIPS)},
  year={2018}
}

@Article{lemeunier:specae,
    author       = {Clément Lemeunier and 
                    Florence Denis and 
                    Guillaume Lavoué and 
                    Florent Dupont},
    title        = {Representation learning of 3D meshes using an Autoencoder in the spectral domain},
    journal      = {Computers \& Graphics},
    year         = {2022}
}

@InProceedings{xu:pointllm,
author="Xu, Runsen
and Wang, Xiaolong
and Wang, Tai
and Chen, Yilun
and Pang, Jiangmiao
and Lin, Dahua",
title="PointLLM: Empowering Large Language Models to Understand Point Clouds",
booktitle="European Conference on Computer Vision (ECCV)",
year="2025",
}

@inproceedings{achlioptas:shapetalk,
title={{ShapeTalk}: A Language Dataset and Framework for 3D Shape Edits and Deformations},
author={Achlioptas, Panos and Huang, Ian and Sung, Minhyuk and Tulyakov, Sergey and Guibas, Leonidas},    
booktitle={Conference on Computer Vision and Pattern Recognition (CVPR)},    
year={2023}}

@Article{kubik:dental-dataset-1,
AUTHOR = {Kubík, Tibor and Španěl, Michal},
TITLE = {LMVSegRNN and Poseidon3D: Addressing Challenging Teeth Segmentation Cases in 3D Dental Surface Orthodontic Scans},
JOURNAL = {Bioengineering},
YEAR = {2024},
}

@article{wang:dental-dataset-2,
  title={A 3D dental model dataset with pre/post-orthodontic treatment for automatic tooth alignment},
  author={Shaofeng Wang and Changsong Lei and Yaqian Liang and Jun Sun and Xianju Xie and Yajie Wang and Feifei Zuo and Yuxin Bai and Song Li and Yongjin Liu},
  journal={Scientific Data},
  year={2024},
}

@InProceedings{tan:dental-dataset-3,
author="Tan, Yuwen
and Xiang, Xiang
and Chen, Yifeng
and Jing, Hongyi
and Ye, Shiyang
and Xue, Chaoran
and Xu, Hui",
title="Coupling Bracket Segmentation and Tooth Surface Reconstruction on 3D Dental Models",
booktitle="Medical Image Computing and Computer Assisted Intervention (MICCAI)",
year="2023",
}

@article{golriz:dental-dataset-4,
title = {Personalized dental crown design: A point-to-mesh completion network},
journal = {Medical Image Analysis (MedIA)},
year = {2025},
author = {Golriz Hosseinimanesh and Ammar Alsheghri and Julia Keren and Farida Cheriet and Francois Guibault},
}

@inproceedings{qi:pointnet,
  title={Pointnet: Deep learning on point sets for 3d classification and segmentation},
  author={Qi, Charles R and Su, Hao and Mo, Kaichun and Guibas, Leonidas J},
  booktitle={Conference on Computer Vision and Pattern Recognition (CVPR)},
  year={2017}
}

@inproceedings{wu:ptv3,
  title={Point transformer v3: Simpler faster stronger},
  author={Wu, Xiaoyang and Jiang, Li and Wang, Peng-Shuai and Liu, Zhijian and Liu, Xihui and Qiao, Yu and Ouyang, Wanli and He, Tong and Zhao, Hengshuang},
  booktitle={Conference on Computer Vision and Pattern Recognition (CVPR)},
  year={2024}
}

@article{reuter:shapeDNA,
  title={Laplace--Beltrami spectra as ‘Shape-DNA’of surfaces and solids},
  author={Reuter, Martin and Wolter, Franz-Erich and Peinecke, Niklas},
  journal={Computer-Aided Design},
  year={2006},
}

@article{reuter:discreteLBO,
  title={Discrete Laplace--Beltrami operators for shape analysis and segmentation},
  author={Reuter, Martin and Biasotti, Silvia and Giorgi, Daniela and Patan{\`e}, Giuseppe and Spagnuolo, Michela},
  journal={Computers \& Graphics},
  year={2009},
}

@article{lemeunier:spectrans,
title = {SpecTrHuMS: Spectral transformer for human mesh sequence learning},
journal = {Computers \& Graphics},
year = {2023},
author = {Clément Lemeunier and Florence Denis and Guillaume Lavoué and Florent Dupont},
}

@article{becker:point-deep-models,
title = {Discriminative and generative models for anatomical shape analysis on point clouds with deep neural networks},
journal = {Medical Image Analysis (MedIA)},
year = {2021},
author = {Benjamín Gutiérrez-Becker and Ignacio Sarasua and Christian Wachinger},
}

@inproceedings{luo:point-diffusion,
  author = {Luo, Shitong and Hu, Wei},
  title = {Diffusion Probabilistic Models for 3D Point Cloud Generation},
  booktitle = {Conference on Computer Vision and Pattern Recognition (CVPR)},
  year = {2021}
}

@inproceedings{naeem:div-metric-cov,
author = {Naeem, Muhammad Ferjad and Oh, Seong Joon and Uh, Youngjung and Choi, Yunjey and Yoo, Jaejun},
title = {Reliable fidelity and diversity metrics for generative models},
year = {2020},
booktitle = {International Conference on Machine Learning (ICML)},
}

@inproceedings{zhu:ptdiff-correspondences,
title={Point-Based Shape Representation Generation with a Correspondence-Preserving Diffusion Model},
author={Shen Zhu and Yinzhu Jin and Ifrah Zawar and Tom Fletcher},
booktitle={Medical Imaging with Deep Learning (MIDL)},
year={2025},
}

@ARTICLE{biffi:explainable-ae,
  author={Biffi, Carlo and Cerrolaza, Juan J. and Tarroni, Giacomo and Bai, Wenjia and de Marvao, Antonio and Oktay, Ozan and Ledig, Christian and Le Folgoc, Loic and Kamnitsas, Konstantinos and Doumou, Georgia and Duan, Jinming and Prasad, Sanjay K. and Cook, Stuart A. and O’Regan, Declan P. and Rueckert, Daniel},
  journal={IEEE Transactions on Medical Imaging}, 
  title={Explainable Anatomical Shape Analysis Through Deep Hierarchical Generative Models}, 
  year={2020},
  volume={39},
  number={6},
  pages={2088-2099},
  doi={10.1109/TMI.2020.2964499}}


\clearpage



\end{document}